\documentclass[sn-mathphys,Numbered]{sn-jnl}

\usepackage{graphicx}%
\usepackage{multirow}%
\usepackage{amsmath,amssymb,amsfonts}%
\usepackage{amsthm}%
\usepackage{mathrsfs}%
\usepackage[title]{appendix}%
\usepackage{xcolor}%
\usepackage{textcomp}%
\usepackage{manyfoot}%
\usepackage{booktabs}%
\usepackage{algorithm}%
\usepackage{algorithmicx}%
\usepackage{algpseudocode}%
\usepackage{listings}%

\newcommand{\reve}[1]{{{\color{black}#1}}}
\newcommand{\revf}[1]{{{\color{black}#1}}}
\newcommand{\revn}[1]{{{\color{black}#1}}}
\newcommand{\rev}[1]{{{\color{black}#1}}}

\begin{document}

\title[Article Title]{Design Principles for Lifelong Learning AI Accelerators}


\author*[1]{\fnm{Dhireesha} \sur{Kudithipudi}}\email{dk@utsa.edu}

\author[1]{\fnm{Anurag} \sur{Daram}}

\author[1]{\fnm{Abdullah M.} \sur{Zyarah}}

\author[1]{\fnm{Fatima Tuz} \sur{Zohora}}
\author[2]{\fnm{James B.} \sur{Aimone}}
\author[3]{\fnm{Angel} \sur{Yanguas-Gil}}
\author[1,4]{\fnm{Nicholas} \sur{Soures}}
\author[5]{\fnm{Emre} \sur{Neftci}}
\author[6]{\fnm{Matthew} \sur{Mattina}}
\author[7]{\fnm{Vincenzo} \sur{Lomonaco}}
\author[8]{\fnm{Clare D.} \sur{Thiem}}
\author[9]{\fnm{Benjamin} \sur{Epstein}}

\affil*[1]{\orgname{University of Texas at San Antonio}, \orgaddress{\city{San Antonio}, \state{TX}, \country{USA}}}

\affil[2]{\orgname{Sandia National Laboratories}, \orgaddress{\city{Albuquerque}, \state{NM}, \country{USA}}}

\affil[3]{\orgname{Argonne National Laboratory}, \orgaddress{\city{Lemont}, \state{IL}, \country{USA}}}

\affil[4]{\orgname{Rochester Institute of Technology}, \orgaddress{\city{Rochester}, \state{NY}, \country{USA}}}

\affil[5]{\orgname{Forschungszentrum J\"ulich and RWTH Aachen}, \orgaddress{ \state{Aachen}, \country{Germany}}}

\affil[6]{\orgname{Tenstorrent Inc.}, \orgaddress{ \state{MA}, \country{USA}}}

\affil[7]{\orgname{University of Pisa}, \orgaddress{ \state{Pisa PI}, \country{Italy}}}

\affil[8]{\orgname{Air Force Research Laboratory}, \orgaddress{ \city{Rome}, \state{NY}, \country{USA}}}

\affil[9]{\orgname{ECS Federal}, \orgaddress{ \city{Arlington}, \state{VA}, \country{USA}}}

\abstract{\reve{Lifelong learning — an agent’s ability to learn throughout its lifetime — is a hallmark of biological learning systems and a central challenge for artificial intelligence (AI). The development of lifelong learning algorithms could lead to a range of novel AI applications, but this will also require the development of appropriate hardware accelerators, particularly if the models are to be deployed on edge platforms, which have strict size, weight, and power constraints. Here, we explore the design of lifelong learning AI accelerators that are intended for deployment in untethered environments. We identify key desirable capabilities for lifelong learning accelerators and highlight metrics to evaluate such accelerators. We then discuss current edge AI accelerators and explore the future design of lifelong learning accelerators, considering the role that different emerging technologies could play.}}

\maketitle

\section*{}\label{sec1}

\reve{Lifelong learning (also known as continual learning) is an approach to artificial intelligence (AI) in which a model is expected to learn from noisy, unpredictable, and changing data distributions while continually consolidating knowledge about new information. The model must transfer previously acquired knowledge forward to new tasks, transfer new knowledge backward to previously learnt tasks, and adapt quickly to contextual changes~\cite{kudithipudi2022biological, delange2021continual}. The capabilities of AI systems have advanced notably in recent years, but designing lifelong learning machines remains difficult. Algorithmic-level innovations are important to address this problem. However, to be of practical value, lifelong learning models must often be deployed on physical hardware at the edge, under strict size, weight, and power constraints. And this will require advances in hardware accelerators for lifelong learning.

Current AI accelerators with on-device learning capabilities support aspects of lifelong learning. (Note, lifelong learning and continual learning are often used interchangeably, but continual learning is not equivalent to online on-device learning.) Edge AI accelerators, in particular, contain limited computational capabilities and operate under battery power, and various challenges — including those related to dataflow, external memory access and computation — remain to be addressed with these systems. The workload profile for lifelong learning also has different characteristics on the edge. These characteristics include being able to process data at variable frequencies while learning relevant features from them, operating under memory and computational constraints, and optimising for energy-accuracy trade-offs in real-time. The characteristics also limit the direct application of optimisation techniques often used in cloud environments. In addition, few of the design approaches provided for edge AI accelerators can transfer to large-scale systems.

In this Perspective, we examine the development of lifelong-learning capable digital hardware accelerators for untethered devices. We first consider fundamental aspects of lifelong learning algorithms and then identify the hardware design requirements that digital accelerators must meet in order to support lifelong learning. We discuss the importance of standardised metrics for lifelong learning systems, and provide an overview of current accelerator designs. We also explore the future of lifelong learning accelerator designs and consider the role that different emerging technologies could play.}

\section*{\rev{Fundamentals of lifelong} learning}
\label{sec:cl_fundamentals}
\reve{The term \textit{lifelong learning} refers to a system's ability to autonomously operate in, interact with, and learn from its environment~\cite{thrun1995lifelong, mccloskey1989catastrophic, McClelland_etal95_whyther_edit}. This requires the system to be able to improve its performance through the acquisition of new knowledge and learning to operate under energy and resource constraints. \textit{More specifically, a lifelong learning system needs to function in dynamic and noisy environments, rapidly adapt to novel situations, minimise forgetting when learning new tasks, transfer information between tasks, and operate without explicit task identification.}

Several learning paradigms and features have been involved in the process of arriving at this definition of lifelong learning. The concept includes transfer learning in which the goal was to recycle/reuse the learnt representations for other tasks~\cite{pratt1991direct}. This was followed by multi-task learning (MTL), which had the goal of improving generalisation by leveraging domain-specific information in related tasks ~\cite{caruana1997multitask}; however, in this setting, tasks are trained jointly and not sequentially. Transfer learning and MTL evolved into few-shot learning ~\cite{fei2006one}, in which the goal is to learn from a limited number of examples with supervised information in the target domain. This drove the approach towards learning to learn (also called meta-learning), where a system learns to optimise the objective on its own; an approach that aims to achieve the rapid, general adaptation that biological brains can offer ~\cite{thrun1998learning}. In tandem, studies on biological brains have shown that there is a combination of learning mechanisms that support lifelong learning, rather than a single one \cite{kudithipudi2022biological}. Drawing from this and more, researchers consider that a number of these learning paradigms constitute a subset of lifelong learning~.

Methods have been devised to address different aspects of lifelong learning. These methods are \textit{synaptic consolidation}, \textit{dynamic architectures}, and \textit{replay} (Fig. 1).}

\begin{figure}
    \centering
    \includegraphics[width = \linewidth]{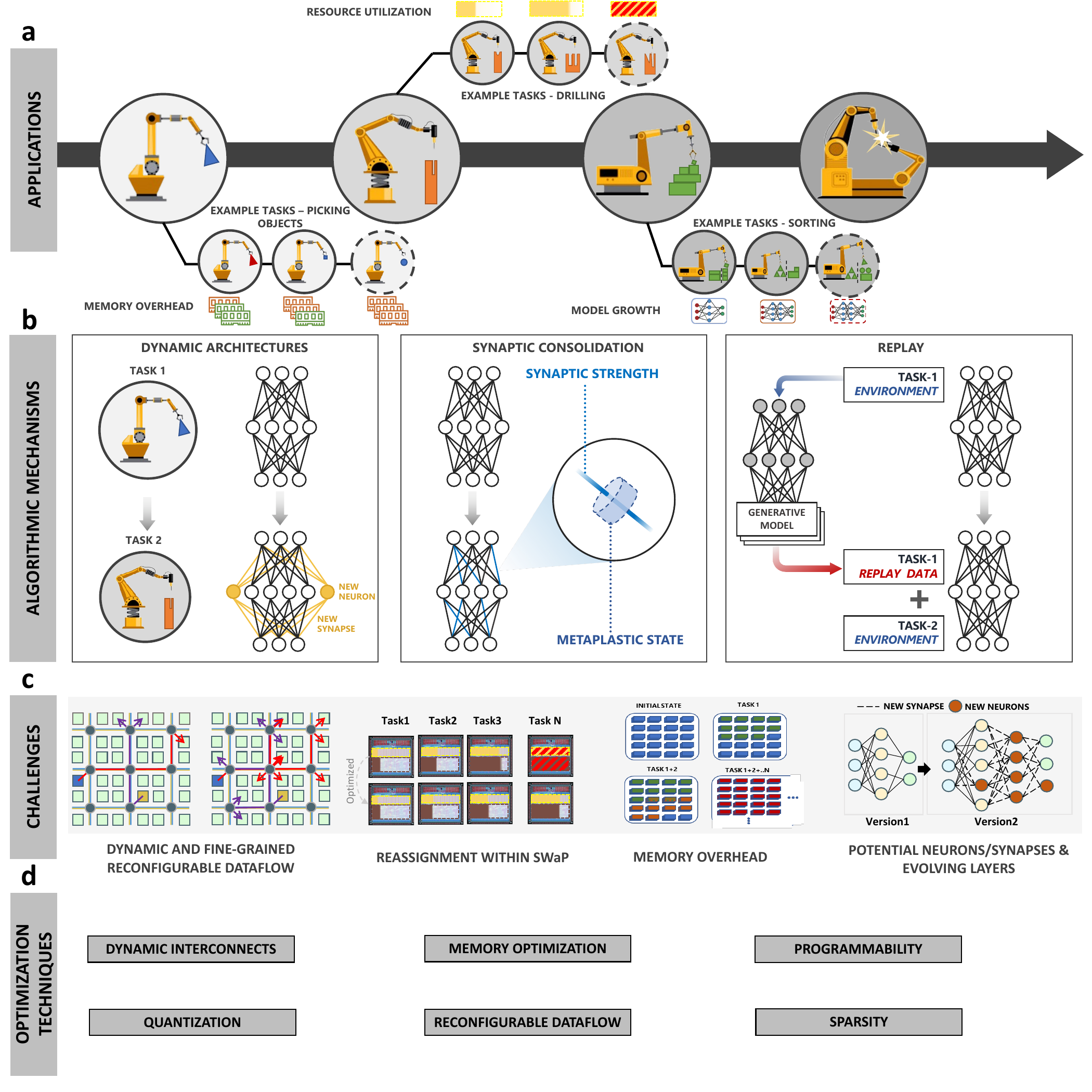}

    \caption{\reve{\textbf{Addressing lifelong learning in AI systems} - \textbf{a}, 
    Applications : Lifelong learning shown in the context of sequential tasks (large circles) and sub-tasks(smaller circles) with varying degrees of similarity, and the associated hardware challenges. 
    \textbf{b}, Algorithmic mechanisms: A broad class of mechanisms that address lifelong learning. Dynamic architectures either add or prune network resources to adapt to the changing environment. Regularization methods restrict the plasticity of synapses to preserve knowledge from the past. Replay methods interleave rehearsal of previous knowledge while learning new tasks. \textbf{c}, Hardware challenges: Lifelong learning imposes new constraints on AI accelerators, such as the ability to reconfigure datapaths at a fine granularity in real-time, dynamically reassign compute and memory resources within a size, weight, and power(SWaP) budget, limit memory overhead for replay buffers, and generate potential synapses, new neurons and layers rapidly. \textbf{d}, Optimization techniques (Bottom): Hardware design challenges can be addressed by performing aggressive optimizations across the design stack. Few examples are dynamic interconnects that are reliable and scalable, quantization to $>$4-bit precisions during training,  hardware programmability, incorporating high bandwidth memory, supporting reconfigurable dataflow and sparsity.}}
    
    \label{fig:overview}
\end{figure}
\reve{\textbf{Synaptic consolidation} is a way to preserve synaptic parameters when learning new tasks. The most common method is to add regularisation terms to the loss function to maintain previous synaptic strengths or neural activations ~\cite{Kirkpatrick_etal17_overcata,zenke2017continual}. Another approach is to use more complex synapse models with memory-preserving mechanisms such as metaplasticity~\cite{Laborieux_etal21}, multiple weight components operating on different timescales~\cite{TACOS1}, or probabilistic synapses~\cite{SS}.

\textbf{Dynamic architectures} expand network capacity in order to solve a particular task, using top-down control of resources. The expansion of network capacity takes different forms, including periodic additions of new neurons or addition of entire new networks or layers dedicated to solving new tasks~\cite{ebrahimi2020adversarial1, pandit2020relational}. Another approach is to dynamically gate the network components according to the task identity. This is often achieved by means of a task oracle (that is, an autoencoder capable of separating specific classes and tasks) or by meta-learning a gating function \cite{masse2018alleviating}.

\textbf{Replay} involves the presentation of samples representative of previously learnt tasks interleaved with samples from the task currently being trained. Replay helps bring the training data closer to being independent and identically distributed, as would be the case if all tasks had been trained jointly. This allows the network to be trained to maximise performance across all tasks (as opposed to only learning the current task) and to learn inter-task boundaries. To replay data in neural networks, previous training samples (or internal representations) are encoded, stored, and recalled from a memory buffer or recreated by a generative model ~\cite{Rebuffi:2016, Lopez-Paz:2017, van2020brain}. Methods are evolving to use more sophisticated algorithms for both the selection of samples to be stored in the replay buffer and for the selection of samples to be replayed from the buffer \cite{hayes2021replay}.

Several machine learning models have been developed to address lifelong learning, with a heavy emphasis on the catastrophic forgetting problem using the methods discussed above, as well as some hybrid models that combine features from two or more of them. Increasingly, AI researchers are taking inspiration from discoveries in neuroscience that help explain how biological organisms are able to learn continually~\cite{kudithipudi2022biological}.}  

\section*{\reve{Key capabilities for lifelong learning accelerators}}
\label{sec:cl_acc_design_features}
\reve{Recent research on synaptic consolidation~\cite{Laborieux_etal21, TACOS1, SS, Kirkpatrick_etal17_overcata}, dynamic architectures~\cite{ebrahimi2020adversarial1, pandit2020relational} and replay methods~\cite{van2020brain, Rebuffi:2016} show promise in addressing aspects of lifelong learning. By studying these methods, we have assembled a list of six desirable capabilities for AI accelerators: on-device learning, resource reassignment within budget, model recoverability, synaptic consolidation, structural plasticity, and replay memory.

The first three capabilities are general ones that apply to any lifelong learning method; the last three are specific to one or more of the lifelong learning methods discussed above. The relevance of each capability to a specific design will depend on the type of accelerator and the target application.

\subsection*{On-device learning} A lifelong learning system needs to continuously learn from non-stationary data distributions over varying time-scales~\cite{mundt2023wholistic}. Since the system has to learn in real-time, on-device learning is crucial to avoid latency associated with data transmission to the cloud. Generally, for on-device learning, design considerations are required i) when batching data of large samples on limited memory resources and ii) when storing and mapping a large pool of intermediate model parameters in compact formats to minimise data movement cost. While translating to the context of lifelong learning, additional challenges arise: optimising complex computations observed in consolidation methods so that the energy cost is reduced and minimising the latency to access previous samples in replay methods.

\subsection*{Resource reassignment within budget} 
Lifelong learning systems must have the ability to reallocate or distribute resources within a size, weight, area, and power (SWaP) budget at runtime, and quite often this could entail different granularities. This suggests that the system must be capable of parametric, neuronal or memory reassignments during its lifetime ~\cite{kwon2021exploring}. This is a challenging requirement — especially when hardware is tightly optimised for one method, model, or functionality. Furthermore,  to allocate resources at fine-granularity one needs to identify points of operation that are pareto-optimal, while ensuring that the architecture remains flexible to different tasks. Additional challenges may arise when the size of the input and output layer differs between tasks~\cite{van2019three,gupta2021continual}, such as accommodating the model expansion on the fly and/or the inability to encode and store data within a fixed size.

\subsection*{Model recoverability} 
One challenge in developing a lifelong learning AI system is to establish confidence in the model, and more so when the system is changing autonomously~\cite{seshia2022toward}. In a software environment, it is possible to checkpoint a model’s state, preserving the overall model or maintaining a history of updates. This tracking provides a valuable record of a model’s past states that can prove essential for either diagnostic purposes or for reverting to a previous state (if an online update led to failure). Several of the design choices that make AI accelerators more efficient make checkpointing a model’s configuration dynamically impractical, if not impossible.

\subsection*{Synaptic consolidation} 
Models incorporating synaptic consolidation typically use multiple internal states to learn at different timescales using several loss functions, probabilistic synapses, reference or target weight values or other synaptic states in addition to magnitude (\textit{i.e.}, metaplastic state, consolidation tag). Each of these consolidation mechanisms entails auxiliary information stored on-device and additional operations performed during any learning process. In general, for consolidation mechanisms, a lifelong learning accelerator needs to store and associate auxiliary information with specific synapses/neurons and support custom loss functions.

\subsection*{Structural plasticity}
 
Structural plasticity relates to physical changes in the model, including the addition/removal of synapses (e.g. synaptogenesis or synaptic pruning) or of neurons (e.g. neurogenesis/neural pruning), gating and attention mechanisms, and mixture of experts~\cite{fernando2017pathnet, lee2020neural, pandit2020relational}. While static allocation of pools of neurons and/or synapses is possible at design time, it is still challenging to model and train such highly dynamic architectures. The underlying accelerators should support fine-grain runtime reconfigurability to add or reallocate a new pool of memory and computational resources. This problem is exacerbated when reallocation has to occur under a limited SWaP budget.

\subsection*{Replay memory} 
Replay mechanisms might require a continuously growing memory as the model learns new tasks. Replay comes in two varieties, known as waking and sleeping replay, both of which need to be supported by accelerators. In waking replay, to prevent forgetting previously learnt tasks, selected samples representative of the earlier tasks are interleaved while training a new task. During sleep replay, the system rehearses only samples from previous tasks to consolidate knowledge. Memory access during sleep replay can be at a slower clock cycle, since the model does not need to respond to real-time stimuli. Memory storage can be off-chip. Waking replay, on the other hand, requires on-chip buffers that can update the system state at a faster rate without disrupting learning on the current task. Overall, memory storage and access patterns for replay mechanisms are of mixed latency and are distributed.

The current generation of accelerators includes some highly programmable devices, but there are not any that support the full set of capabilities listed above. And those that do support some of the features~\cite{teslap100, norrie2021design} do not meet the size, weight, area, and power (SWaP) budget constraints imposed by untethered devices.
}

\section*{Initial metrics to evaluate \rev{lifelong} learning accelerators}
\label{sec:metrics}

\reve{In lifelong learning scenarios, the statistical properties of the input stream cannot be assumed to be stationary, an assumption that underlies traditional statistical learning theory approaches. This limits the usability of standard evaluation protocols and metrics described in the machine learning literature. However, the methods used to assess the quality and capability of lifelong learning systems have been evolving along with the progress in models and applications ~\cite{lifelong_metrics}.

Table~\ref{tab:accelerator_metrics} lists recent lifelong learning metrics for evaluating algorithms. Based on pilot implementations~\cite{zohora2021metaplasticnet, SCOLAR}, we also propose additional metrics for lifelong learning accelerators (communication overhead and multi-tenancy).

A common benchmark for assessing how a system learns a sequence of tasks is to measure the mean accuracy across all tasks after the model has been trained on each task at least once (that is, at the end of the overall training sequence). This metric can be compared to the accuracy achieved when all tasks are trained jointly to obtain a measure of the amount of forgetting that is due to sequential learning. However, there are nuances that are not reflected in this type of measurement. For example, it does not distinguish between a system that learns the first task perfectly but is unable to learn subsequent ones and a system that learns the last task perfectly but forgets the preceding ones. For this reason, studies~\cite{Lopez-Paz:2017, diaz2018don} that measure how much performance changes in prior tasks (backward transfer), and how performance changes in downstream tasks (forward transfer) have provided more insight into how systems address continual learning and where their limitations lie.

\begin{sidewaystable}\tiny
\caption{Overview of current metrics for \rev{lifelong} learning algorithms~\cite{lifelong_metrics} and proposed metrics for the accelerators.}\label{tab:accelerator_metrics}

\begin{tabular}{c|c|c}
\toprule%
\textbf{Current Metric} & \textbf{Formula} & \textbf{Metric Assessment of System}\\
\midrule
Mean Accuracy (MA)\footnotemark[1] & $\mathrm{MA} = \sum\limits_{t=1}^{N} \frac{\mathrm{R}^{t,N}}{N}$ & \shortstack[l]{Average performance of model on all the tasks experienced} \\ \hline
    Memory Overhead (MO)\footnotemark[2] & $\mathrm{MO} = min\Big(1, \frac{1}{N}\sum\limits_{i=1}^{N}\frac{Mem(\theta_i)}{Mem(\theta_b)}\Big)$ & \shortstack[l]{Average overhead in memory a model requires per task} \\ \hline
    Forward Transfer (FWT) & \reve{$FWT = \frac{1}{N-1}\sum\limits_{k=1}^{N-1}\sum\limits_{t=k+1}^{N} \frac{\mathrm{R}^{t,k}-R^{t,k-1}}{N-k}$ }& \shortstack[l]{\revn{Average change in accuracy across all tasks $t > k$, after learning task $T^k$}} \\ \hline
    Backwards Transfer (BWT) & \reve{$BWT = \frac{2}{N(N-1)}\sum\limits_{k=2}^{N}\sum\limits_{t=1}^{k-1} \mathrm{R}^{t,k} - \mathrm{R}^{t,k-1}$} &  \shortstack[l]{\revn{Average change in accuracy on tasks $t < k$, after learning task $T^k$}}\\ \hline
    
    Performance Recovery (PR)\footnotemark[3] & $PR = \frac{d}{dt} (T_{recovery}(t))$ & \revn{Slope of recovery times of model in response to a change}\\ \hline
    Performance Maintenance (PM) &\reve{ $PM = \frac{1}{N-1}\sum\limits_{k=1}^{N-1}   \sum\limits_{t=k+1}^N \frac{(\mathrm{R}^{k,t} - \mathrm{R}^{k,k})}{N-k}$} & \shortstack[l]{\revn{Average change in performance on a task after learning new tasks}}\\ \hline
    Sample Efficiency (SE)\footnotemark[4] & $SE = \frac{\mathrm{R}^{t,t}}{b^{t}} \times \frac{T_{sat}^{b^t}}{T_{sat}^{\mathrm{R}^{t,t}}}$ & \shortstack[l]{\revn{Efficiency of a lifelong learner vs a single task expert to reach saturation in performance}}\\ \hline
    
    \rev{Arithmetic Intensity\footnotemark[5]} & $\frac{OPs}{Byte}$ & \shortstack[l]{\revn{Reuse efficiency of accelerator as number of operations per byte of memory traffic}} \\ \hline
    
    \rev{Energy Efficiency} & $\frac{OPs}{W}$ & \shortstack[l]{\revn{Efficiency of the system as a ratio} of computing throughput of the system per W}\\ \hline

    Learning Cost & $N_{trainsteps} \times N_{updates} \times \frac{Energy\_Cost}{Update}$ & Energy cost for the training process of the accelerator \\ \hline
    Area Efficiency & $\frac{OPs}{mm^2}$ & The number of operations per each $mm^2$ of the chip for a given technology node \\ \hline
    \rev{Working Memory Footprint}&  $Bytes$& \revn{Net} memory size required for learning different tasks\\ \hline

    \rev{Communication Overhead\footnotemark[6,*]}& $f(D,C)$ & \revn{\shortstack[l]{Cost of communication as a function of the distance between two memory accesses (D) \\ and associated memory access cost (C)}}\\ \hline
    
      \rev{Multi-Tenancy\footnotemark[7,*]}&\reve{$\Delta({T_{(t+1)},T_{t}+L_{t}})$} &  Time lapse between the runtime of two sequential tasks\\ 

\botrule
\end{tabular}

\footnotetext[1]{$\mathrm{R}^{t,k}$ represents the accuracy of the \rev{lifelong} learning on task number $t$ after learning task number $k$. $N$ represents the total number of tasks the model experiences with $k,t \leq N$}
\footnotetext[2]{$\theta_i$ represents the average amount of memory a model requires per task, and $\theta_b$ baseline model's memory size.}
\footnotetext[3]{$T_{recovery}(t)$ is the function of the curve formed by the recovery times for the \rev{lifelong} learning model. Recovery time \\ measures the time to recover its performance when a change is observed.}
\footnotetext[4]{\rev{$b^t$ represents the performance of a single task expert model on a task $t$. $T_{sat}^{b^t}$ and $T_{sat}^{\mathrm{R}^{t,t}}$ denote time to performance saturation  \\ ($T$ can be represented by number of samples to reach saturation) for single task expert and lifelong learner, respectively. \\ SE measures the ratio of time (number of samples) to reach saturation scaled up by the ratio of peak performance for a \\ single task expert and the lifelong learning model.} }
\footnotetext[5]{\rev{$OPs$ refers to the number of compute operations required for performing a task and \textit{Byte} refers to the number of bytes \\ accessed in memory. This provides the ratio of number of operations per byte of memory traffic.}}
\footnotetext[6]{\rev{D refers to the distance of the data in memory hierarchy and C refers to the cost of memory access.}}
\footnotetext[7]{\rev{Latency of the task L = $\frac{instructions}{task}\times\frac{cycles}{instruction}\times\frac{seconds}{cycle}$, $T_{t}$ is the time at which task t starts.}}
\footnotetext[*]{\revf{Represent the newly proposed hardware metrics, which are important for evaluating lifelong learning accelerators.}}

\end{sidewaystable}

Other metrics specific to lifelong learning include more granular methods for measuring trade-offs between plasticity and stability, measuring how continual learning can be leveraged to learn faster or improve performance compared to a baseline model (by using prior knowledge), measuring the time to recover performance after task transitions (performance recovery), and measuring performance degradation on previous tasks after each new task is learnt (performance maintenance). Beyond performance, it is also important to study models in terms of applicability~\cite{lesort2020continual}; this includes metrics quantifying how robust models are to noise, failures and data ordering, and autonomy of the model (for example, does it need supervision, task oracles).

Another important dimension of evaluation, especially for edge AI accelerators, is sample efficiency and scalability. These aspects can be quantified in terms of memory overhead, training speed, and network growth over time (which should preferably be bounded irrespective of the amount of data processed). Furthermore, for the accelerators, we consider the cost of data movement and reuse for all the methods (arithmetic intensity), energy efficiency of the system, total memory footprint that can increase with lifelong learning methods, the communication overhead that is associated with accessing data from higher-order memory for replay or plasticity, and multi-tenancy capturing the systems response rate when executing sequential tasks for real-time operations.

In addition, there are metrics and evaluation protocols specific to accelerator design. For example, metrics that highlight the efficiency-efficacy trade-off, method tunability as it depends on specific application requirements~\cite{ravaglia2020memory}, performance variability as a function of data sequence length and out-of-distribution streams, and continual evaluation regimes that measure worst-case performance, necessary when deploying critical real-world applications~\cite{de2023continual}. Several metrics and benchmarking platforms have been published for AI inference accelerators~\cite{reddi2020mlperf, vanschoren2014openml} and lifelong learning algorithms ~\cite{lifelong_metrics}, but there is a need to identify and develop new benchmarks and workloads suitable for continual learning accelerators on a larger scale, including hardware and algorithmic metrics.}

\section*{Lifelong learning on current untethered AI accelerators}
\label{sec:acc_landscape}

\reve{AI accelerators can be categorised based on those that support traditional rate-based implementations and those that support spiking neural networks. Tables \ref{tab:accelerator_low_power} and \ref{tab:accelerator_low_power_SNN} provide an overview of digital AI chips (rate-based and spiking) that can perform on-device learning in untethered environments. We split the accelerators into two tables ~\cite{davies2019benchmarks} because the baseline resolution of the computations and associated metrics such as performance (TOP/s) and power are different for spiking accelerators and rate-based accelerators and because SNN algorithms generally support different network topologies and have different encoding schemes. We also note there are differences in the choice of design optimisations for the two sets of accelerators.

Traditional accelerators for rate-based AI algorithms have primarily been focused on implementing and optimising DNNs, CNNs, and RNNs, generally trained using gradient descent applied with backpropagation. Design choices during the development of these accelerators focus on microarchitectural exploration~\cite{jouppi2020domain, chen2019eyeriss}, energy-efficient memory hierarchies [49, 50, 51], flexible dataflow distribution ~\cite{gu2020dlux, lee20197, han2020df}, domain-specific compute optimisations such as quantisation, pruning, and compression ~\cite{evolver, HNPU, kimetal}, and hardware-software co-design techniques~\cite{brainwave}. More recently, a rate-based accelerator with latent replay and on-device continual learning has been proposed~\cite{ravaglia2020memory}. The design leverages the benefits of quantisation and tiling to reduce compute and memory overheads.

Compared to rate-based accelerators, spiking neuromorphic accelerators are designed to support algorithmic models that more closely mimic the functionality of their biological counterparts. This often involves more complex synaptic and neuronal dynamics, which are inherently temporal in nature. The typical characteristic of a spiking neuron is how it integrates information over time, and only releases a spike once the accumulated information crosses a threshold. Neuromorphic systems tailored for SNNs have demonstrated efficiency in a number of ways: low cost of a single spike, asynchronous and sparse communication, and cheaper synaptic operations that do not require multiplication. Though these features can be achieved in non-spiking domains, unlike recurrent neural networks, SNNs have an inherent temporal aspect in the neuron dynamics without the need for recurrent connections while offering greater applicability and computational power than binary neural networks.

In the context of lifelong learning, several promising methods draw inspiration from neural plasticity, including spike-timing-dependent plasticity, heterosynaptic weight decay and consolidation, and neuromodulation~\cite{kudithipudi2022biological}. The bio-plausible learning models generally have local learning, neuronal and synaptic plasticity rules that require fine granular control on the hardware substrates. Neuromorphic accelerators inherently offer a high degree of freedom to optimise dynamically at such fine granularity compared to the rate-based accelerators~\cite{davies2018loihi}. For example, triplet STDP rules~\cite{pfister2006triplets} that learn temporal correlations between pre- and post-synaptic spike neurons (useful for rapid adaptation) are more amenable for deployment on neuromorphic accelerators that support integrate-fire dynamics.

The current spiking accelerators capable of learning can be divided into two broad categories: large-scale spiking accelerators\cite{davies2018loihi, furber2014spinnaker, demler2019brainchip} and accelerators targeted towards edge platforms \cite{nguyen2021review, frenkel2022reckon, frenkel20180, chen20184096, dean2016vlsi, Chicca_etal13_neurelec_edit}. Large-scale spiking accelerators support a wider range of applications or cortical simulations that focus on scalability and programmability~\cite{davies2018loihi, furber2014spinnaker, demler2019brainchip}. These accelerators employ multiple independent parallel cores to realise the neurons and synapses of the network. The cores are connected through network-on-chip interconnects, which offer greater flexibility with high-bandwidth inter-chip or inter-board interfaces. In contrast, accelerators targeted towards edge platforms consist of a single core that supports various degrees of network connectivity, such as full connectivity in a crossbar architecture or locally-competitive- algorithm based architectures for sparse inputs~\cite{basu2022spiking}.

Although most accelerators in both categories usually incorporate local unsupervised learning for on-device training, recent efforts are advancing toward incorporating multilayer spiking neural networks with supervised training \cite{frenkel2022reckon}. There have also been extensions of supervised SNN models to lifelong learning. For example, a digital spiking network was designed for lifelong learning tasks using surrogate gradient-based training~\cite{SCOLAR}. This accelerator uses activity-dependent metaplasticity to enable lifelong learning. For example, memory access overhead is reduced by using sparse spike-based communication, where only indices of neurons with active spikes are transmitted. Quantisation and compute-lite linear metaplasticity functions reduce memory and computational overhead.}

\begin{figure}
    \centering
    \includegraphics[width=0.98\textwidth]{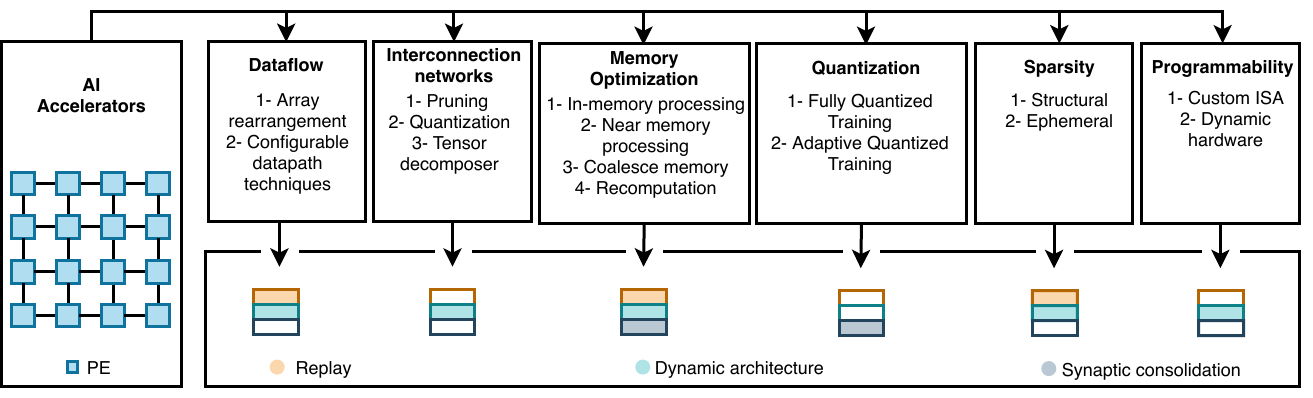}
    \caption{\reve{\textbf{Hardware optimisations for lifelong learning in AI accelerators.} The bar plots indicate which lifelong learning features are affected by each optimisation technique. }
}
    \label{fig:mapping_features}
\end{figure}

\begin{sidewaystable}\tiny
\caption{Overview of AI chips  with on-device training and feature optimisations that can support a \revf{subset of the} features of lifelong learning in untethered environments.}\label{tab:accelerator_low_power}%
\begin{tabular}{c|c|c|c|c|c|c|c|c}
\toprule
\textbf{Chip} & \textbf{Quantization} & \textbf{\shortstack{Neural \\Network(s)}} & \textbf{Power (W)}  & \textbf{Performance\footnotemark{7}} & \textbf{\shortstack{On-chip \\Memory\footnotemark{6}}} & \textbf{Energy Efficiency\footnotemark{7}} & \textbf{Sparsity} & \textbf{Dataflow} \\

\midrule
 PuDianNao~\cite{liu2015pudiannao} & FP16/FP32\footnotemark[1] & \shortstack{7 ML \\algorithms\footnotemark[2]}& \shortstack{596 mW\\ (65nm)}&  1.056 TOP/s & 32 kB&  1.77 TOPS/W & - & - \\ \hline 
        PNPU~\cite{kim2020146_56}  & FP8, FP16 & DNN, CNN & \revf{\shortstack{425 mW@1.1V\\ (65nm)}} & \revf{\shortstack{0.31 TFLOP/s \\(FP16)\\ 0.61 TFLOP/s \\(FP8)}} & \revf{338 kB} & \revf{\shortstack{0.72 
        TFLOPS/W \\(FP16)\\1.44 TFLOPS/W \\(FP8)}} & \shortstack{In/W/Out \\zero skipping\footnotemark[3]}& - \\ \hline 
        GANPU~\cite{kang20207_54}  & FP8, FP16 & GAN & \revf{\shortstack{647 mW@1.1V \\(65nm)}} & \revf{\shortstack{0.54 TFLOPS \\(FP16)\\1.08 TFLOPS\\ (FP8)}} & \revf{676 kB}& \revf{\shortstack{0.83 TFLOPS/W \\(FP16)\\1.66 TFLOPS/W\\ (FP8)}} &  \shortstack{In/Out \\zero skipping} & \revf{\shortstack{Reconfigurable 2D \\ mesh connection}} \\ \hline
        Evolver~\cite{evolver}  & \shortstack{INT2, INT4, \\INT8}  & \revf{DNN, CNN} & \shortstack{36 mW@1.1V \\(28nm)} & \revf{\shortstack{2.195 TOP/s \\(INT2$\times2$)\\0.137 TOP/s \\(INT8$\times8$)}} & \revf{416 kB} & \revf{\shortstack{173 TOPS/W\\(INT2)\\32.9 TOPS/W \\(INT8)}} &  \shortstack{In/Out \\zero skipping} & \revf{\shortstack{Tile-wise dataflow \\reconfiguration}} \\ \hline
        HNPU~\cite{HNPU}&  SDFXP 4/8/12/16\tnote{4} & DNN, CNN & \shortstack{1162 mW\\ (28nm)} &  \shortstack{3.07 TOP/s \\(SDFXP4)} & \revf{552 kB}  &  \revf{~50.3 
         TOPS/W (FXP4)} & \shortstack{In-/Out-slice \\zero skipping} & - \\ \hline
        LNPU~\cite{lee20197}& FGMP FP8-FP16\tnote{5}& \shortstack{DNN, CNN, \\RNN} & \shortstack{367 mW \\(65nm)} &  \shortstack{$>$0.6 TOP/s \\(FP8)} & \revf{372 kB}  & \revf{\shortstack{1.74 TOPS/W (16b)\\3.48 TOPS/W (8b)}}& \shortstack{In-zero \\skipping} & \shortstack{Reversible datapath \\ Tiled weight rearrangement}\\ \hline
        DF-LNPU~\cite{han2020df} & FXP13/16, FP8 & \shortstack{MLP, CNN,\\ RNN} & \shortstack{\revf{424 mW$@$1.1V}\\ (65nm)} & \revf{\shortstack{LC:0.151 TOP/S\footnotemark[7]\\ZCC:0.3-1 TOP/s\footnotemark[7]}} & \revf{337 kB} &\revf{\shortstack{LC:0.61-1.1 
         TOPS/W\footnotemark[7]  \\ ZCC:1.7 TOPS/W\footnotemark[7]}}& \shortstack{In/W \\zero-skipping} & \shortstack{Transpose in\\ custom SRAM} \\ \hline
        Agrawal et al.~\cite{agrawal20217nm}& Hybrid-FP8, FP16 & \shortstack{MLP, CNN,\\ RNN} & n/a (7nm) &  16 TFLOPS & & \revf{0.98 TOPS/W }&- & Flexible Datapath MUXs\\ \hline
        SOVC18~\cite{fleischer2018scalable_46}& FP16 & \shortstack{MLP, CNN, \\LSTM} & n/a (14nm) & \revf{1.5 TFLOP/s} & \revf{2 MB}& - &- & 2D Torus \\ \hline
        SSCL20 ~\cite{yinetal}& \revf{FXP16} & \revf{CNN} & \shortstack{299 mW$@$1V\\ (65nm)} & 0.15 TOP/s & \revf{1.12 MB}& \revf{0.5 TOPS/W}&- & \shortstack{Diagonal storage \\pattern  with \\bit rotator}\\ \hline
        ISSCC19 ~\cite{kimetal}& \shortstack{bfloat16,\\ FXP16/8/4}& \shortstack{MLP, CNN, \\RNN}& \shortstack{\revf{196 
        mW$@$1.1V}\\ (65nm)} & 0.204 TOP/s & \revf{\shortstack{139 MB/20000 \\ experiences\footnotemark[6]}}& \revf{\shortstack{2.16 TFLOPS/W \\(16b)}}&- & \shortstack{Transposable PE\\ array datapath}\\ \hline
        \revf{CHIMERA ~\cite{prabhu2022chimera}} & INT 8 & DNN, CNN & \shortstack{418 mW \\(40nm)} & \revf{0.92 TOP/s} & \revf{2 MB\footnotemark[6]} &\revf{2.2 TOPS/W} & Gradient sparsity & Weight Stationary \\ \hline
        {\shortstack{Tenstorrent \\Wormhole \\\cite{wormhole, vasiljevic2021compute}}} & \shortstack{FP16, FP8,\\ bfloat16} & \shortstack{DNN, CNN,\\ LSTM} & \shortstack{80 W\\ (12nm)} & \revf{430 TOP/S} & \revf{120 MB} &-& \shortstack{Activation and\\  parameter sparsity}& - \\ \hline
         \rev{SIGMA~\cite{qin2020sigma}} & FP16/32 & \shortstack{DNN, RNN\\, CNN} & \shortstack{22.3 W \\(28nm)} & 10.8 TFLOPS &\revf{68 MB} &\revf{0.48 TFLOPS/W }&\shortstack{Bitmap \\compression} & \shortstack{Reduction tree\\ microarchitecture}\\ 
\botrule
\end{tabular}

\footnotetext[1]{Multiplication in FP16, accumulation uses FP32}
\footnotetext[2]{ K-means, k-nearest neighbours, naive bayes, support vector machine, linear regression, classification tree \\and deep neural network.}
\footnotetext[3]{In/W/Out refer to Input/Weight/Output}
\footnotetext[4]{SDFXP - Stochastic dynamic fixed point representation}
\footnotetext[5]{FGMP - Fine-grained mixed precision}
\footnotetext[6]{All AI chips are using on-chip SRAM except \cite{prabhu2022chimera} which uses RRAM. In case of~\cite{kimetal}, it is not reported.}
\footnotetext[7]{Energy efficiency and performance with no sparsity. In~\cite{han2020df} ZCC refers to zero-skip convolution cores and LC is the learning core.}
\end{sidewaystable}

\begin{sidewaystable}\tiny
\caption{Overview of spiking neural network chips with on-device training and feature optimizations that can support a subset of the features of \reve{lifelong} learning in untethered environments.}\label{tab:accelerator_low_power_SNN}%
\begin{tabular}{c|c|c|c|c|c|c|c}
\toprule
\textbf{Chip} & \textbf{Quantisation} & \textbf{Neural Network(s)} & \textbf{Power}  & \textbf{Throughput} & \revf{\textbf{On-chip Memory \tnote{8}}} &\textbf{Connectivity} & \textbf{Sparsity} \\
\midrule
Loihi~\cite{davies2018loihi, shrestha2021hardware} & INT1-INT9 \tnote{1}  & \revf{Spiking MLP, Spiking } & \revf{420 mW (14 nm)} & 50 FPS (10kHz) \tnote{2}& \revf{33 MB } & NoC & \revf{ Sparse activity-dependent} \\  
         &   & \revf{ CNN, LSM} &  &  &  &  & \revf{ asynchronous flow control} \\ \hline   

\revf{SpiNNaker 2~\cite{mayr2019spinnaker, hoppner2018spinnaker2}}  & \revf{FXP32, FP32} & \revf{DNN, SNN} & \revf{$\sim$ 0.72 W ( 22nm) \tnote{3}} & \revf{4.6 TOP/s (250MHz)} & \revf{18MB SRAM \tnote{4} } & \revf{NoC} & \revf{DVFS based on}  \\ 
           &  & \revf{CNN} &  &  & &  & \revf{ input sparsity}  \\\hline 
BrainChip  & INT1, INT2 & \revf{CNN, SNN} & 434 mW (80 FPS, 28nm) & 1.5 TOP/s (300MHz) & \revf{8 MB} & NoC & \revf{Sparse event-driven} \\ 
        Akida~\cite{demler2019brainchip} &  INT4  & &  &  &  & & \revf{ computations} \\\hline
        
        ODIN \cite{frenkel20180} & INT4  & \revf{SNN (SDSP)} \tnote{5} & 477 $\mu$W (28nm)& 37.5 MSOP/s (75MHz) \tnote{6}&  \revf{36 kB (SRAM)}& AER bus & \revf{Sparse event-driven} \\
         &  &  & & &  &  & \revf{computation} \\
        
         \hline
        Intel SNN Chip \cite{chen20184096} & INT7  & \revf{SNN (STDP)} & \revf{6.2 mW (inference, 10nm)}& 25.2 GSOP/s \tnote{6}& \revf{896 kB} & AER NoC & \revf{Input/weight-based skipping} \\ \hline
        {ReckON ~\cite{frenkel2022reckon}} & INT8  & \revf{Spiking RNN} & 114-150 $\mu$W(28nm) &  - & \revf{138 kB (SRAM)} & AER bus & \revf{ET/STE-based skipping\tnote{7}} \\ \hline
        
        DANNA~\cite{dean2016vlsi} & INT8 & \revf{SNN (STDP)} &  n/a (130nm) & -- & --- & Nearest neighbor & \revf{--- } \\\hline  
        SCOLAR~\cite{SCOLAR} & FxP16 & \revf{SNN}  &  21.25 mW(65nm) &  250 MOP/s (10MHz)& \revf{100 kB (SRAM)} & NoC & \revf{Sparse event-driven }\\
        &  &   &   &  &  & & \revf{ computation}\\
\botrule
\end{tabular}

\footnotetext[1]{Signed or unsigned integer or mixed-precision number is supported.}
\footnotetext[2]{FPS - Frames per second.}
\footnotetext[3]{Processing Element (PE) power.}
\footnotetext[4]{Spinnaker 2 is also equipped with 8GB off-chip DRAM.}
\footnotetext[5]{SDSP - Spike-driven Synaptic Plasticity.}
\footnotetext[6]{SOP - Synaptic Operations, MSOP - Mega Synaptic Operations, GSOP - Giga Synaptic Operations.}
\footnotetext[7]{ET- Eligibility trace, STE - Straight-through estimator of spiking activation function.}
\footnotetext[8]{All chips use SRAM as on-chip memory.}
\end{sidewaystable}

\subsection*{Optimisation techniques}\label{sec:optimization_techniques}
\reve{There are a number of optimisation techniques used in current AI accelerators: reconfigurable dataflow, memory optimisation, dynamic interconnection networks, quantisation, sparsity, and programmability. For each of these techniques, we suggest extensions that would facilitate lifelong learning (Fig. 2). We also highlight how each method may impact the metrics described in Table \ref{tab:accelerator_metrics}. However, it is important to note that the optimisation techniques listed can help to realise only a subset of features for lifelong learning.}

\subsubsection*{\reve{Reconfigurable dataflow}}  On-device training requires iterative processing to compute optimal model parameters. The training procedure typically consists of three stages, forward pass, backward pass and parameter update. \rev{The three stages share a processing element (PE) array but have different dataflows, which leads to different memory access patterns for the same data array. In addition, the optimal memory layout is different for each data flow, making optimising the operations of all three training stages difficult. This incongruity between dataflow and memory layout causes redundant memory access operations (each memory access consumes $\sim$200$\times$ more energy than compute units~\cite{chen2017using}) and reduces core utilisation, resulting in speed and ~\textit{energy efficiency} degradation. Recent studies have described several techniques to address this problem, which can be broadly classified into array rearrangement and configurable dataflow techniques. Array rearrangement techniques by performing matrix transpose in custom SRAM, introducing diagonal storage pattern~\cite{yinetal} or tiled weight rearrangement~\cite{lee20197, lu20192_50} are some examples that are currently used, and are presented in Table \ref{tab:accelerator_low_power}. Configurable dataflows~\cite{fleischer2018scalable_46, jouppi2020domain} include reversing dataflow for weights and outputs~\cite{lee20197}, or exchanging paths between weights and inputs~\cite{kimetal}.  
}

\rev{When considering lifelong learning accelerators, novel dataflow optimisation techniques can play a large role in handling structural plasticity and performing efficient replay. Unlike the three stages observed in on-device learning using backpropagation, structural plasticity involves a dynamic change in network architecture, requiring different optimal memory layouts for each change. Similarly, for replay, the system transitions from learning from streaming data to accessing, processing, and learning from batched samples of previous experiences. Such dynamic behaviour significantly increases the search space for finding the mapping (translating the model into its hardware-compatible representation) that best optimises data movement for improved arithmetic intensity. Additionally, to accommodate this flexibility in structure, some studies have explored techniques such as reduction tree microachitectures, atomic dataflow using graph-level scheduling and mapping~\cite{qin2020sigma, giannoula2022sparsep} to handle sparse and irregular data movement, but trade speed for higher power consumption. The aforementioned techniques can be useful in supporting mechanisms such as episodic replay, pruning and dynamic gating; however, further complex architectural changes like neurogenesis or the addition of entirely new layers or networks in real-time pose challenges in adopting high degrees of reconfigurability and identification of optimal mapping space without impacting \textit{energy efficiency} of the accelerator.  
}

\subsubsection*{\reve{Memory optimisation}}  
\rev{
Off-chip memory access can account for more than 80\% of total energy consumption at inference \cite{li2018smartshuttle}. Consequently, during training the overhead is more pronounced, since calculating gradients \revn{can require} up to 10$\times$ as much memory access as updating the weights~\cite{jouppi2020domain}. Therefore, it is crucial to reduce energy consumption due to memory access for energy-constrained platforms. One way to achieve this goal is to devise techniques to reduce the cost of individual memory access. The major component of energy consumption during memory access operation is due to data communication rather than data access; the former can incur up to $\sim99\%$ of total energy consumption \cite{dally2022model}. This has motivated designers to reduce the communication distance between memory and processing. A key technique in this regard is processing in memory which performs computation within the memory array. A similar technique is near-memory computation, which brings compute cores closer to memory. \revn{DLUX \cite{gu2020dlux} which is an accelerator that leverages near-bank computation in 3D memory has shown on average $42 \times$ improvement in \textit{energy efficiency} on representative data centre training workloads compared to the Tesla V100 GPU.} \revf{ BrainChip's Akida \cite{demler2019brainchip} and Intel's SNN chip \cite{chen20184096} adopt near-memory computation with distributed compute cores where each core is assigned dedicated on-chip memory.}

Another way to reduce energy consumption due to memory access is to reduce the number of accesses altogether. To this end, dataflow and data layout in memory can be optimised to maximally coalesce memory access and reuse data \cite{lee20197}. Another technique is to use recomputation to save memory access \cite{chen2016training}.} In this technique, the intermediate states necessary for the backward pass are recomputed from the checkpoint rather than stored in memory. These are some of the prominent techniques that help tackle the \textit{memory overhead} for on-device training.

\begin{figure}
    \centering
    \includegraphics[width=1.0\textwidth]{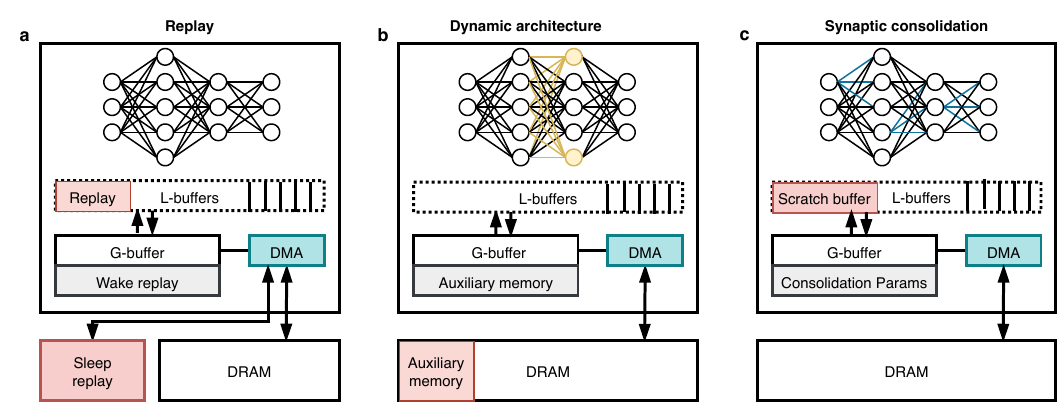}
    \caption{\reve{\textbf{Memory models}. Potential memory models to enable lifelong learning features in AI accelerators, for methods such as: replay (\textbf{a}), structural plasticity (\textbf{b}), and synaptic consolidation (\textbf{c}). Heterogeneous and adaptive memory architectures that support data access with variable latency, high bandwidth, flexible data storage with minimal multi-tenancy (time lapse between two sequential tasks) will play an important role in these accelerators.}
    }
    \label{fig:memory_hierarchy}
\end{figure}

\rev{Lifelong learning methods that mitigate catastrophic forgetting naturally exacerbate \textit{memory overhead} during training. A conceptual memory organisation for the three lifelong learning methods is shown in Fig. 3.
Structural plasticity and synaptic consolidation require additional storage on-device for the network parameters and structural pointers.}
In structural plasticity, the number of weights, activations, and gradient information that needs to be stored will grow over a system's lifetime, \rev{ hence the \textit{memory overhead} can linearly increase with the number of tasks encountered in the models lifetime \cite{de2021continual}}). In Fig. 3, we show this overhead with additional auxiliary memory that can be kept in sleep mode when not in use.
Synaptic consolidation often involves using additional states to represent information such as metaplasticity, consolidated weights, or probabilistic information. \rev{This can lead to up to $\sim 3.5 \times$ \textit{memory overhead} for a short sequence of tasks \cite{TACOS1}. Moreover, consolidation approaches may require scratchpad memory to temporarily store network activity or gradient information to determine the importance of parameters. Lastly, in replay methods, in addition to storing network information, there may be a need to continuously store prior experiences in some form of memory buffer or at least some form of knowledge representing prior experiences from which synthetic data can be generated. It can lead to up to $2 \times$ memory storage overhead in addition to the buffers \cite{de2021continual}, which may increase with the addition of newer tasks leading to unbounded memory requirements. The optimum size of the replay buffer is usually chosen based on the performance in the problem being solved and the resource constraints. While simpler tasks may require storing a small percentage of data (1\%), more complex tasks may need more storage for acceptable performance \cite{merlin2022practical}. Depending on the sample size, frequency of access and the stage of replay (sleep or wake), this data can be located in higher levels of memory hierarchy or in buffers that are close to the compute substrate, as shown in Fig. 3. In practical scenarios, a combination of methods will be chosen that can compound the \textit{memory overhead}. It is worth emphasising that the cost of the overhead is largely determined by the location of the data in the memory hierarchy, rather than its amount. Since fetching data from an off-chip DRAM can incur} $\sim 33\times$ more energy compared to on-chip global buffer \cite{chen2017using}, even a small overhead can incur substantial energy cost as it resides in higher levels of memory. 
Consequently, memory optimisation is critical for most existing lifelong learning algorithms on untethered devices.

\subsubsection*{\reve{Dynamic interconnection networks}} \rev{The interconnect topology and the speed of the link can critically impact the flexibility, functionality, and real-time performance of an accelerator. Topologies such as mesh, tree, ring, and hybrid are commonly used to support global connectivity, especially in AI accelerators, where common patterns are observed in weight updates and underlying communication. The selection of topology and packet-based communication mode (unicast, multi-cast, or broadcast) directly depends on factors such as latency, peak and average bandwidth, and traffic patterns. Common NoC topologies are variants of the H-Tree, hierarchical tree, and toroidal mesh~\revf{\cite{kang20207_54, HNPU, davies2018loihi, furber2014spinnaker}}. The 2D torus topology offers high throughput as it enables equal bisection bandwidth between the two halves of the network~\cite{jouppi2020domain,fleischer2018scalable_46, mayr2019spinnaker} and supports parallel synchronous and asynchronous stochastic gradient descent operations. \revf{Additional schemes such as population-based hierarchical connectivity introduced in Loihi~\cite{davies2018loihi}, can reduce the connectivity resources by an order of magnitude by utilising predefined connectivity templates mapped to specific populations.} }

\revn{To support features such as neuronal pruning and neurogenesis which are dynamically triggered by novel input stimuli, it is important to select a topology that is reliable, scalable, and offers fine-grained routing reconfigurability.}
 \revn{Rapid reallocation or generation of new resources is required during high periods of activity. Full connectivity to all the processing elements is desirable, as new neurons can be generated in any layer. At the same time, greater routing feasibility is preferred at the local nodes as potential synapses are formed. Furthermore, fast router links for global and local connectivity will be required to facilitate real-time updates.}

\rev{It is important to note that support for the above mentioned features comes at the cost of increased node count and, therefore, higher power and area - undesirable for untethered devices. For instance, full connectivity leads to large-area and poor scalability of the models. \revn{On the other hand, fast router links such as high bandwidth metallic
interconnects and advanced digital routers are power hungry (for example, 2GHz on-chip network router 
fabricated at 16nm CMOS process consumes 2.4 to 2.8W of power~\cite{nedbailo2022designing}). Optical interconnects or hybrid optical interconnects (2D or 3D topologies) can be potential solutions in the long term ~\cite{bashir2019survey, shastri2021photonics}.}}

\begin{figure}
    \centering
    \includegraphics[width=0.8\textwidth]{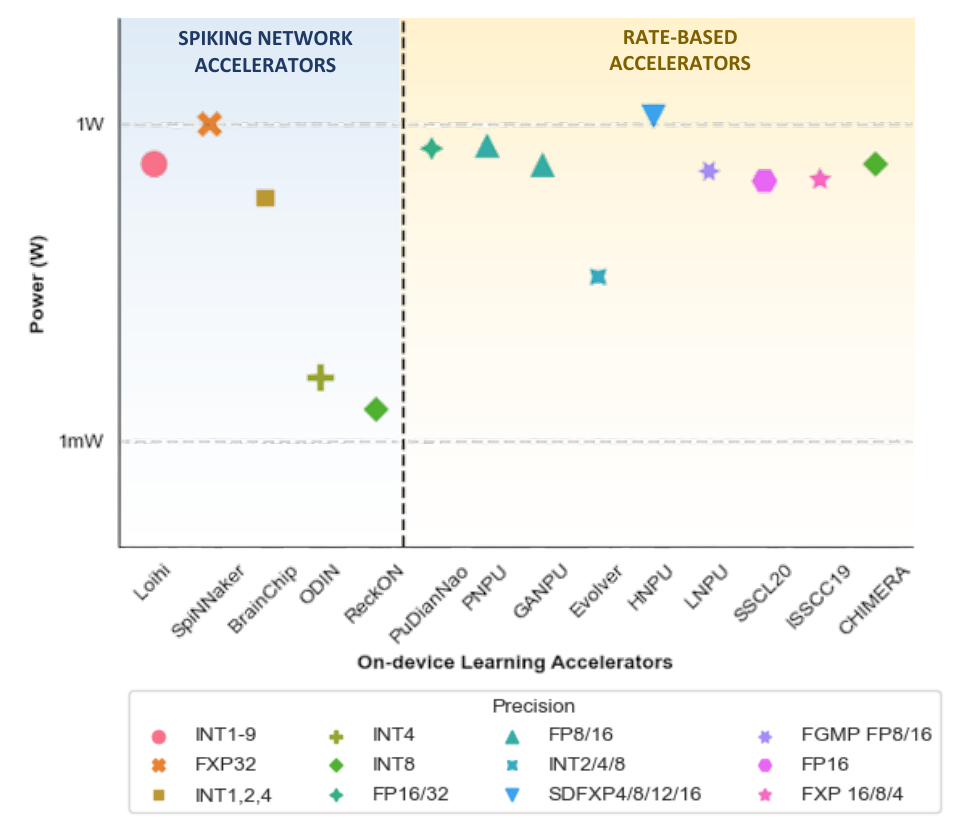}
    \caption{
  \reve{\textbf{Maximum accelerator power with respect to quantisation and numerical formats.} In general, accelerators operating on lower bit precision ($<8$-bit) demonstrate lower power consumption.}}

    \label{fig:accelerator_quantization}
\end{figure}

\subsubsection*{\reve{Quantisation}} \rev{ Quantisation can offer significant benefits when operating under resource constraints, enabling the reassignment of resources and increasing the feasibility of model check-pointing. 
For example, quantising model parameters from 32-bit to 8-bit, can 
reduce the model size by 50\%\textendash75\% and reduce the energy consumption by 67\%\textendash75\% ~\cite{krishnamoorthi2020}. Fig. 4 and \revf{Tables \ref{tab:accelerator_low_power} and ~\ref{tab:accelerator_low_power_SNN} show how different accelerators use quantised precision to achieve low energy and high throughput~\cite{HNPU, kim2020146_56, kang20207_54,agrawal20217nm,lee20197,davies2018loihi, frenkel2022reckon, dean2016vlsi}}. \revf{However, also note that few of the accelerators shown in Fig. 4 demonstrate significantly lower power due to other factors such as chip size and on-chip memory (refer to Tables \ref{tab:accelerator_low_power} and ~\ref{tab:accelerator_low_power_SNN}) for details.} 

Existing approaches for minimisation of quantisation error fall under three categories: quantisation-aware training, post-training quantisation, and fully quantised training. Quantisation-conscious training approaches train or fine-tune models with simulated quantisation operations to learn quantised weights and activations. On the other hand, no training is involved in the post-training quantisation approach; the high-precision parameters are directly converted to low-precision parameters. While the former two approaches are used for inference applications, fully quantised training can be used for training a model that quantises gradients in addition to weights and activations \cite{chen2020statistical}.
\rev{Quantisation on hardware generally involves reduction of bit-width and designing custom processing elements to operate the optimised bit-precision. For example, studies have implemented a 2x{fp} 8/1x{fp}16 (fp - floating point) configurable fused-multiply add~\cite{lee20197}; some techniques removed {fp} special values such as \textit{NaN, Inf} and \textit{subnormal} during training stage~\cite{oh20203_53}, and some newer works suggest near-memory-processing techniques~\cite{kim2021gradpim, zhao2021cambricon} wherein simple operations are performed by data with high-bit precision in external memory and the data is fetched to the accelerator after quantisation in the memory. }}

Lifelong learning models often require high-precision computation and are challenging to operate in a low-precision regime. For example, in regularisation techniques, previous knowledge is protected by reducing the changes to the weights that are deemed important. However, low-precision weights prohibit such precise changes in magnitude, thus requiring longer duration to learn a task. Another challenge is that these models require multiple quantisation levels for the different lifelong learning methods. When these models are deployed on accelerators incorporating energy-efficient MACs, the challenge exacerbates due to the dynamic quantisation requirements on-device. 
Therefore, the system should have knowledge of the statistical parameters to identify the right quantisation mechanism. However, these mechanisms have a large search space to narrow down to an optimal quantisation level.
Studies suggest that quantisation can be used to improve model fit on-device and throughput, where these systems carry out continuous online training with training frequencies from \textit{ an hourly to daily} basis and training duration between \textit{seconds to days}~\cite{hazelwood2018applied}. 
Other techniques, such as adaptive quantised training ~\cite{yao2021hawq, zhao2021distribution}, 
show promise in their applicability to lifelong learning accelerators. We can assess the goodness of a quantisation approach by performing ablation studies to determine how byte size affects the \textit{ arithmetic intensity} and thereby throughput for real-time learning. Additional insights can be gained by studying the \textit{memory footprint} for a specific type of quantisation.

   \subsubsection*{\reve{Sparsity}} \rev{Recent studies on ML models show that sparsification can lead to a reduction in
model size by $1$\textendash$2$ orders of magnitude~\cite{hoefler2021sparsity}.  In sparsification methods, we reduce representational complexity using only a subset of the features (eliminating ineffectual computations and reducing storage by encoding only nonzero values) and achieve the corresponding improvements in computational, storage and \textit{energy efficiency} without loss of accuracy. Unlike mainstream over-parameterized models that tend to overfit to the data and degrade generalisation to unseen examples, sparsity can help with generalisation. Sparsification can be either structural (\textit{i.e.} weights, neurons, heads) or ephemeral (\textit{i.e.} activations, gradients, errors). Generally, accelerators aim for either sparse matrix-vector or sparse matrix-matrix multiplications. Sparsity is inherently unstructured (\textit{e.g.}, activations, gradients), where the \textit{nonzero} elements are randomly scattered. Often to improve hardware execution, the structure is introduced through model operators or weight pruning in coarse-grain blocks. Most accelerators utilise co-design approaches of a sparse training algorithm and hardware. Specific approaches include structural weight sparsity by storing weights in a compressed block format, supporting the arbitrary reuse of matrices or their elements, \revf{sparse compression with event-driven activity gating, using address event representation} and using blocked bitmap storage to implement sparse vector products~\cite{qin2020sigma, lee20197, HNPU, evolver, giannoula2022sparsep, davies2018loihi, frenkel2022reckon}. A sparse matrix storage format determines ranges of sparsity where
it performs most efficiently. \revf{Additionally, spiking accelerators incorporate the aforementioned optimisations since SNNs fundamentally feature a high degree of sparseness in their activity~\cite{furber2014spinnaker, davies2018loihi, frenkel20180}}. Tables \ref{tab:accelerator_low_power} and ~\ref{tab:accelerator_low_power_SNN} show how different accelerators handle sparse computations.}

\rev{Sparsity can be quite unstructured and fine-grained in lifelong learning methods. Dynamic sparsity techniques that iteratively combine pruning and regrowth of elements during the training process, as in~\cite{zyarah2019neuromorphic, pandit2020relational}, will be more suitable than static sparsity approaches that do not account for model structure updates during training. Techniques such as ephemeral sparsity during training, sparsification during training, and dully sparse training can be utilised based on the method. Structural plasticity methods introduce load imbalance depending on the distribution of zeros across different processing elements (\textit{i.e.} inter-PE and intra-PE load imbalances). This requires a dedicated block that can share the workload at run-time and support asynchronous execution. It seems that the optimisation space for these methods is quite large. Algorithms can guide the types of sparsity that can be supported. A process like neurogenesis can be valuable for sparse high-dimensional representations that remain stable with adaptation~\cite{9395703}. Several of the hardware metrics such as \textit{memory footprint, learning cost, power efficiency} are directly impacted by the level of sparsity, extraction of non-zeros, data decoding, and synchronisation. Furthermore, \textit{communication overhead} captures the peak utilisation of accelerators' computational resources and off-chip bandwidth.}

\subsubsection*{\reve{Programmability}} \rev{Any physical accelerator is designed within fixed} resource and energy budgets. \rev{Programmability allows a system to perform flexible operations within a fixed budget. However,} an inverse relationship is observed between the programmability and \textit{energy efficiency} of the accelerators~\cite{nowatzki2016pushing}. Prior studies have focused on improving the programmability of the accelerators to support multiple learning algorithms and architectures \rev{within a power budget of $<1W$~\cite{liu2015pudiannao,chen2020survey,davies2018loihi}}. For example, researchers have explored different techniques such as using custom instruction set architectures~\cite{brainwave, mayr2019spinnaker}, multi-core architectures (large core counts) utilising smart interconnects with each core capable of executing completely distinct programs~\cite{jia2019dissecting} and support for dynamic hardware reconfiguration~\cite{fleischer2018scalable_46, kang20207_54, evolver, dyhard}.

To perform \rev{lifelong} learning in the wild (in untethered scenarios)~\cite{mundt2023wholistic}, \rev{systems may have changing goals and functionalities, utilise different learning rules, or even different models and architectures}. Translating this to accelerator design requires flexibility in selection of programmable methods to update a model's architecture and parameters. Such programmability is crucial for reassigning resources under SWAP constraints where several features are necessary but need not be active simultaneously. For example, a system could periodically switch between employing structural plasticity or complex synapses and replay. In this manner, the advantages of different techniques can be utilised based on the needs at run-time. Additionally, within the paradigm of \rev{lifelong} learning, highly programmable accelerators are expected to support algorithms that evolve over time, enabling efficient implementations of structural plasticity.

\subsection*{\reve{General design considerations}} \revn{ Following Occam's razor, designing lifelong learning accelerators can be seen as a form of multi-scale optimisation and may effectively increase the system robustness and reconfigurability. The greatest difficulty lies in activating all of the above features to their full potential in a single accelerator, so that it can be used for lifelong learning.

\revn{
While lifelong learning accelerators feature the real-time learning capabilities of on-device learning accelerators, the former require quite a few additional features. In particular, lifelong learning accelerators must optimise mapping of sparse and irregular data movement within a limited power budget. Concretely, this optimisation supports structural changes during the model lifecycle. As far as memory is considered, a major difference is in the dynamic memory structures. Memory can grow or shrink at multiple abstractions (\textit{e.g.}, temporary buffers, replay buffers, plasticity modules) based on the lifelong learning mechanism. On the other hand, on-device learning accelerators can experience model changes, but the memory access patterns might not change across inference or training. Reconfigurability is a key element in any accelerator, but for lifelong learning systems, a more fine-granular reconfigurability is necessary to direct resources when rapid plasticity occurs at run-time. 
In recent times, most machine learning models use compression techniques such as sparsity and quantisation. What makes it difficult in the context of lifelong learning is that the early stage structure adaptation and later stage adaptation can be varied, or there can be unstructured sparsity caused by neurogenesis. This means the system has to perform adaptive compression, whether it is quantisation or other forms of compression. New tapered precision numerical formats that are effective during training and inference might come to rescue~\cite{gustafson2017posit, langroudi2021alps}. A critical point is that all of these methods should address the extreme SWaP requirements associated with untethered applications.
}

In general, the surge in different algorithmic mechanisms and the AI models, and evolving evaluation methods cause a widespread in the accelerator designs. This makes it hard to determine the state-of-the-art and make direct comparisons on performance and order-of-magnitude improvements. For example, if ~\cite{SCOLAR} serves as a reference for lifelong learning accelerators, one can state the goal for the accelerators is to operate at 500 MOP/s throughput and $\sim$ 10-15 ms of training latency per sample with 14mW power at 14nm node. 
However, it is important to note that ~\cite{SCOLAR} currently supports regularisation-based mechanisms and is also not rigorously optimised. This might not then be a fair baseline for accelerators that support all three mechanisms or a different class of mechanisms~\cite{slda_ll, zhang2022xst}, as the architectures can lean towards compute-bound or memory-bound based on the algorithmic mechanism.}\revn{We intend to revisit the discussion of lifelong learning on spiking neural network accelerators as they mature.}

\section*{\reve{Future designs for lifelong learning accelerators}}
\label{sec:opportunities}
\begin{figure}
    \centering
    \includegraphics[width=0.8\textwidth]{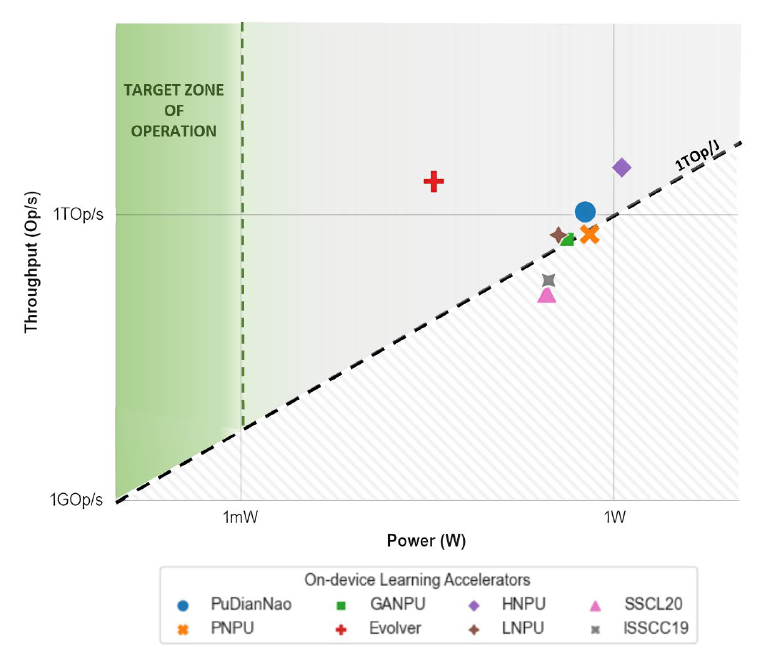}
    \caption{\reve{
    \textbf{Accelerator power vs. throughput.} We observe the power vs. throughput for accelerators 
    that support on-device training in untethered environments  ~\cite{liu2015pudiannao,kim2020146_56,kang20207_54,evolver, HNPU, lee20197, yinetal, kimetal}. The axes are set in log scale. The green region represents the target zone of operation, ($<1mW$ and $>1GOp/s$ ), for lifelong learning edge accelerators \revn{\cite{gao2022energy}}. }}
    \label{fig:accelerator}
\end{figure}
\reve{The tinyML community~\cite{warden2019tinyml} has specified a set of targets for edge AI accelerators: power consumption $<$1mW, numerical precision of $\leq$4-bits, expected best case latency in the range 1-10ns, SRAM capacity of 100kB-150kB, and DRAM capacity of 0.5MB-1MB. Fig. 5 illustrates the power and throughput values of a subset of current AI accelerators, with on-device learning capabilities \cite{liu2015pudiannao,kim2020146_56,kang20207_54,evolver, HNPU, kimetal, lee20197, yinetal}. As can be seen, we still have a ways to go before reaching the targets set for inference accelerators. For instance, the power consumption of most of the current on-device learning accelerators ranges from 36mW to 1.16W, throughput ranges from 0.15 TOP/s to 3 TOP/s, and numerical precision is between 2 and 16-bits (integer and floating point) albeit a couple of exceptions [ Tables.\ref{tab:accelerator_low_power} \& \ref{tab:accelerator_low_power_SNN}].

Simultaneously satisfying all TinyML design criteria may be challenging in the near future, without architectural innovations, aggressive optimisations, and new circuit designs. This challenge is further exacerbated for lifelong learning machines. For a lifelong learning accelerator, at the edge, we propose that the target zone of operation is $<$ 1mW and $>1GOp/s$, which is similar to the neuromorphic system counterparts. Previous studies have identified this zone of operation for the extreme edge ~\cite{gao2022energy}. This zone is suitable for lifelong learning applications in untethered environments, given operational constraints. It should be noted that this zone is an initial target and can be adapted as there is more development in the models and accelerators. Although optimisations in current digital accelerators can extend their capabilities, it is important to rethink accelerator architectures for lifelong learning.}

\subsection*{\rev{Reconfigurable architectures}}
\revn{High degree of reconfigurability is required primarily for structural plasticity mechanisms, as it supports the physical changes in the model primitives. To provide such fidelity, design aspects to consider include : i) a shared bus with flexible bandwidth to interact amongst all layers (\textit{e.g.}- neurogenesis/pruning); ii) presence of dynamic precision accumulators or integrators to circumvent rapid saturation and excessive neuron firing; and iii)  distributed memory (SRAMs), that can be tuned to different sleep modes when unused for timing-sensitive routines (2MB SRAM fabricated in 65nm process can result in power saving of $\sim5000\times$ when running in deep-sleep mode as compared to active mode~\cite{sram_power}}). \revn{Communication schemes in which the synaptic connections are virtually formed or pruned, such as address-event representation (AER)~\cite{mahowald1992vlsi}, enhanced AER~\cite{goldberg2001probabilistic}, or synthetic synapse representation (SSR)~\cite{zyarah2020neuromorphic}, are potential pathways to realise a shared bus with flexible bandwidth.  New tapered numerical formats, such as Posit~\cite{carmichael2019deep}, can be used in the service of dynamic precision modulators. These formats have shown potential to operate with high accuracy and at $>$8-bit precision in both inference and training operations~\cite{langroudi2021alps, murillo2021plam}.}

To address reconfigurability  
in cases where the parameters of a cluster of neurons are regularly tuned (\textit{e.g.}: metaplasticity), time multiplexing~\cite{zyarah2017resource} can be considered. One approach is for the processing elements (PEs) to be time multiplexed   
such that each PE serves multiple neurons, namely one-serve-multiple. This can facilitate neurogenesis or synaptogenesis within preset computational resources. In this approach, additional memory is required to account for the new connections created during structural plasticity. \revf{Moreover, the memory overhead problem is aggravated due to the requirement to store neurons with multiple internal states} \rev{(\textit{e.g.}-sigmoid/ReLU as opposed to spiking neurons/LSTM cells). \revn{In other scenarios where fixed neurons are added, the memory constraint is no longer present and PEs can be equipped with linear-feedback shift registers (LFSRs) to formulate synaptic pathways (as in SSR) or generate the synaptic weights~\cite{zyarah2016optimized}}. Rather than storing synaptic weights, generating them can improve \textit{energy efficiency} and give the accelerator an additional degree of freedom to create or prune synaptic connections. 
When forming and pruning synapses, the model must be aware of the importance of a synaptic connection to the previously learnt information.

\subsection*{Memory topologies}
\rev{ For memory design, there are two main design considerations : i) high data bandwidth for rapid learning and ii) large memory footprint.}

Addressing the first design consideration is particularly challenging, as data transfer across the chip can incur high energy (up to 62\% of total energy in a mobile device \cite{mutlu2023modern}), which limits the bandwidth that can be supported in energy-constrained platforms. \revn{We envision memory organisation biased towards local computation as a promising approach to address this issue. Although current processing-in-memory architectures leverage this principle, these allow minimal modification in memory arrays, so it may prove difficult to accommodate complex computations. Developments to support such computations can prove impactful to realise energy-efficient lifelong learning accelerators. In this regard, processing-near-memory architectures have more potential, since the logic layer being close to memory can offer more flexibility.
In addition, advances in memory technology are also necessary to support high bandwidth data transfer in energy-constrained platforms.} Newer technologies, such as optical interconnects, can be incorporated in 3D memory to improve memory bandwidth and \textit{energy efficiency}.
More fine-grained DRAM architectures are also being explored that activate a smaller portion of the array and reduce the communication distance between I/O and cell array for \textit{energy efficiency} \cite{o2017fine, olgun2022sectored}. Progress on this front can also prove beneficial in supporting high memory bandwidth in lifelong learning accelerators.

The second design consideration of \revn{large} memory footprint motivates us to look toward software-controlled heterogeneous memory for lifelong learning accelerators. In traditional hardware-managed memory hierarchy, high-speed memory incurs a large chip area without contributing to memory capacity, which is not feasible in area-constrained platforms. This issue can be avoided with software-controlled memory. It can optimally use heterogeneous memory to enhance on-chip memory capacity by allocating data with different lifetimes and access frequencies to different types of memory. For example, the gradient/ network activity information required in consolidation mechanisms can be stored in scratchpad memory or any memory technology with low access time and higher endurance while network parameters that are updated on a slower timescale, for example, parameter importance metrics, can be stored in a compact slower memory. Replay methods may require storing a substantial number of samples from previous tasks, but since these samples are usually interleaved with current task samples, their access may be less latency-sensitive. A more compact memory technology with higher access latency such as HBM or HMC could be more suitable for storing them. With advances in emerging memory technologies (more on this in subsection Beyond CMOS and emerging technologies), designers can choose from an array of devices from memory to storage to address the \textit{memory overhead} associated with lifelong learning methods. However, it remains to be understood how memory should be distributed to optimally support data access patterns and \textit{memory overhead} of the different lifelong learning methods. }

{}

\subsection*{On-chip communication}

\revn{In the context of lifelong learning, the Network on Chip (NoC) should be able to meet the varying data and bandwidth requirements without degradation in throughput, as the network grows in real-time and its structure is altered. On-chip communication should arguably promote reliability and availability. 
Time-division-multiplexed communication (supporting asynchronous and synchronous solutions) ~\cite{indiveriintegration, manohar2022hardware}, and dense 3D networks can offer interim solutions to enable multiple tiers of interconnections for structural plasticity mechanisms and regularisation techniques with high bandwidth communication. It is important to note that the lack of high bandwidth on-chip communication has a direct impact on the performance of lifelong learning accelerators, such as slow adaptation for neurogenesis,  throughput limitations for replay methods, missing critical information during asynchronous updates, and executing time-critical updates in regularisation while learning.} 

\revn{Other approaches to improve on-chip communication is to use intra-layer optical interconnects, where a single transmitter delivers data to multiple arbitrarily arranged receivers to support local learning}
\cite{shastri2021photonics}. \rev{Optical wiring and modulators can enable direct photonic broadcast (many to many, without a need for time-multiplexed communication) with minimum latency and energy consumption~\cite{shastri2021photonics}, or point-to-point communication via optical superchannels \revn{(point-to-point communication via optical superchannels uses upto 1Tb/s electro-optical modulator and can offer $\sim$13$\times$ higher bandwidth than advanced digital routers)~\cite{rossi2021electro, nedbailo2022designing}.}}

\subsection*{Beyond CMOS and emerging technologies}

\rev{When we contemplate the potential of technologies beyond CMOS and those that are emerging, for lifelong learning, we can separate them into two distinct categories.

The first category relates to how the inherent characteristics of emerging technologies that are close to maturation can address the computational challenges of lifelong learning. \textit{Memory overhead} is one of the primary
examples: lifelong learning methods incur significant \textit{memory overhead} and should be supported within the form factor of untethered devices. These methods also require memory with a range of latency and endurance criteria to optimise performance.
 
Emerging non-volatile memory (NVM) such as resistive random access memory (RRAM), phase change memory (PCRAM), spin transfer torque magnetoresistive random access memory (STT-MRAM), hold promise in this regard.
RRAM and PCRAM are suitable for meeting memory density requirements, as they are capable of 3D integration, which offers ~3$\times$ higher density compared to contemporary DRAM \cite{yu2022semiconductor}. STT-MRAM devices can be used to store parameters that are updated frequently, as they exhibit fast read and write time and high endurance \cite{park2012future}. Combined with CMOS memory technology, emerging memory devices support heterogeneous architectures with a range of latency, density, and endurance characteristics that can meet the requirements of lifelong learning methods. Mature RRAM and PCM devices can also be incorporated into processing-in-memory (PIM) architectures to support streaming input scenarios. 
It is projected that RRAM-based PIM architecture consumes ~$\sim 57\%$ less energy than SRAM \cite{yu2021rram}. However, device engineering is required to make them operate at lower technology nodes.
Moreover, the devices exhibit non-ideal characteristics such as nonlinearity in tuning, low precision, and limited endurance, which can negatively impact the model performance. These challenges call for further research to improve device characteristics along with creative adaptation of device variability in algorithms and architectures that alleviate the effects of non-idealities.

 The second category relates to how the needs of lifelong learning architectures open up the design space of emerging devices and materials. For example, device properties, such as the ability to modulate the change in internal state at different voltage or current levels, can be leveraged to implement metaplasticity mechanisms \cite{zhu2017emulation, zohora2020metaplasticity}.  Likewise, a performance analysis of memristive devices for online learning has shown that tolerances in the write error of $\approx$40\% can be achieved due to the self-correcting nature of supervised learning algorithms and synaptic plasticity models \cite{ayg2019}. This potentially opens up the design space to devices that may exhibit a lower degree of precision or control of resistive states. 
 Devices with different degrees of retention can contribute to lifelong learning systems: high-retention devices can be used to emulate long-term network parameters, whereas low-retention devices can be leveraged for short-term metaplasticity. Frequent updates may require devices with higher endurance in both scenarios.

It is important to highlight the scarcity of studies that evaluate the potential of emerging technologies in the context of lifelong learning~\cite{zohora2020metaplasticity}. A fundamental question to be explored is what requirements these devices should have to maximise their potential for lifelong learning. In fact, it is
plausible that different parts of an architecture may require different device characteristics.} All these factors open opportunities for research on device, architecture, and algorithm co-design, to leverage beyond CMOS technologies effectively for lifelong learning on untethered platforms.

{}

\section*{\reve{Outlook}}
\label{sec:discussion}

\reve{For AI models to learn efficiently and continually in practical settings, careful consideration of the underlying AI hardware accelerators is required. Rapid innovations are occurring in both lifelong learning models and the underlying hardware. Therefore, model and hardware efficiencies are intertwined, and co-design approaches are essential in order to design good machines. To facilitate co-design between model and hardware, we have outlined features for lifelong learning accelerators that are agnostic or tied to specific methods. These methods serve as the backbone for most existing lifelong learning models. However, as model complexity increases, transitioning from current AI accelerators to the next generation of accelerators will be necessary for both advances in research and practical applications.

We also explored existing accelerators for learning at the edge, and highlighted how lifelong learning can exacerbate the gap between an ideal accelerator and existing AI accelerators. We examined how common optimisation approaches can be adapted to focus on features necessary for lifelong learning, providing a route for existing AI accelerators to become applicable to lifelong learning. Often these optimisations will be application specific depending on the models tried and tested in their corresponding domains prior to deployment. To design a more general-purpose lifelong learning accelerator, the system should have the ability to autonomously self-tune at a finer granularity based on the application, and have the capability to evaluate the trade-offs across the learning features. However, these insights are limited to the scope of existing accelerators, and we also discussed areas that have room for further exploration, particularly beyond CMOS and emerging technologies.

What then is the optimal solution for lifelong learning accelerators? As a starting point, this can include achieving on-device training for high-dimensional and large-scale problems on the cheap, fine granular run-time reprogrammability of compute/memory resources and novel heterogeneous memory architectures that support high-density storage and fast access with increased bandwidth. Moreover, all of these features should be supported in sub-milliWatt power budget so that they can operate seamlessly in untethered environments. We also provided an initial set of metrics that can help assess the different solutions in terms of performance and cost. Although there is unlikely to be one unified metric to clearly identify if one solution is better than another, the metrics discussed should offer the granularity to compare accelerators based on applications.

The brain’s ability to adapt throughout its lifetime makes it an attractive source of inspiration for lifelong learning. However, the complexity of biological learning can be overwhelming. Therefore, it can be helpful to identify key high-level neuroscience insights when considering future directions for the development of lifelong learning accelerators.

First, the brain has a considerable diversity of learning mechanisms spanning a range of spatial and temporal scales and the specific types of plasticity can vary considerably between brain regions and between different types of computational tasks. For example, while a structural plasticity mechanism such as neurogenesis may be useful for domains with continuously evolving input and output dimensionalities, such as episodic memory, it may not be necessary for systems where the input and output dimensionalities are well defined, such as motor control or vision processing \cite{aimone2011resolving}.

Second, related to the task specificity of the learning mechanisms, the brain often dynamically modifies the plasticity itself through neuromodulation or feedback circuits. The use of such top-down mechanisms may control computational processing according to different behavioural contexts, but it can also select which plasticity mechanisms are active in the system.

Third, sleep/wake cycles enable the system to switch between encoding and consolidation modes, and permit the transfer of information between brain regions. Such modulatory control allows a system to differentially regulate what information is learnt.

Furthermore, when considering the future development of lifelong learning accelerators, a number of key challenges and open questions remain. To start, how do we co-design hardware architectures to achieve high degrees of run-time reconfigurability? And what are the associated trade-offs between recovery time, cost, energy, and mean accuracy? Can new architectures support structural changes at fine granularity in order to realise potential synapses? What is the relationship between synaptic plasticity, structural plasticity, learning dynamics, and generalisation? How should memory be distributed among different technologies or topologies to reduce the memory overhead of lifelong learning? How best to use emerging devices with various ranges of read-write latency/energy and endurance to optimally support the memory requirements of lifelong learning? And what should be the ideal distribution between existing and emerging memory technologies?}

\bmhead{Acknowledgments}

We thank members of the Neuromorphic AI lab, Peter Helfer, Tej Pandit, Vedant Karia, and S. Hamed Fatemi Langroudi, for  discussions on this topic. We also thank the reviewers for their valuable feedback. 
Part of this material is based on research sponsored by Air Force Research Laboratory under agreement number FA8750-20-2-1003 through BAA FA8750-19-S-7010. The U.S. Government is authorized to reproduce and distribute reprints for Governmental purposes notwithstanding any copyright notation thereon. The views and conclusions contained herein are those of the authors and should not be interpreted as necessarily representing the official policies or endorsements, either expressed or implied, of the Air Force Research Laboratory or the U.S. Government. This article has been approved for public release; distribution unlimited (Case No. AFRL-2023-3120, 28 Jun 2023). AY acknowledges
Laboratory Directed Research and Development (LDRD) funding from Argonne National Laboratory, provided
by the Director, Office of Science, of the U.S. Department of Energy under Contract No. DE-AC02-06CH11357.

\section*{Declarations}
\bmhead{Competing interests}
The authors declare the following competing interests: The authors declare no competing interests.
\bmhead{Author Contributions}
DK led the team on the design and concept for the manuscript. DK, AD, AZ, FZ, JA, AYG, NS, EN, MM, VL, CT and BE had multiple rounds of discussions in conceptualization of the manuscript and have contributed to the iterative draft manuscripts and the main manuscript text. AD and AZ prepared the figures, with input from DK, FZ, and, NS. AD, FZ and AZ collected data and prepared the tables in the manuscript. DK revised it critically for important intellectual content. All authors commented on the manuscript and reviewed the final-version of the manuscript.

\clearpage
\newpage

\end{document}